\documentclass[10pt,twocolumn,letterpaper]{article}

\usepackage{iccv}              
\usepackage{subcaption}
\usepackage{lipsum}
\usepackage{multirow}
\usepackage{graphicx}
\usepackage{booktabs}
\usepackage{xcolor}
\usepackage{amsmath}
\usepackage{array}
\usepackage{multirow}
\usepackage{color}
\usepackage{colortbl}
\usepackage{framed}
\usepackage{bm}
\usepackage{bbm}
\usepackage{xspace}
\usepackage{enumitem}
\usepackage{xparse}
\usepackage{algorithm}
\usepackage{listings}
\usepackage{url}
\usepackage{wrapfig}
\usepackage{pifont}
\usepackage{tcolorbox}
\usepackage{fontawesome5}
\usepackage[accsupp]{axessibility}

\newcommand{\titledmodelname}{LLaVA-MORE\xspace\includegraphics[width=1.35em]{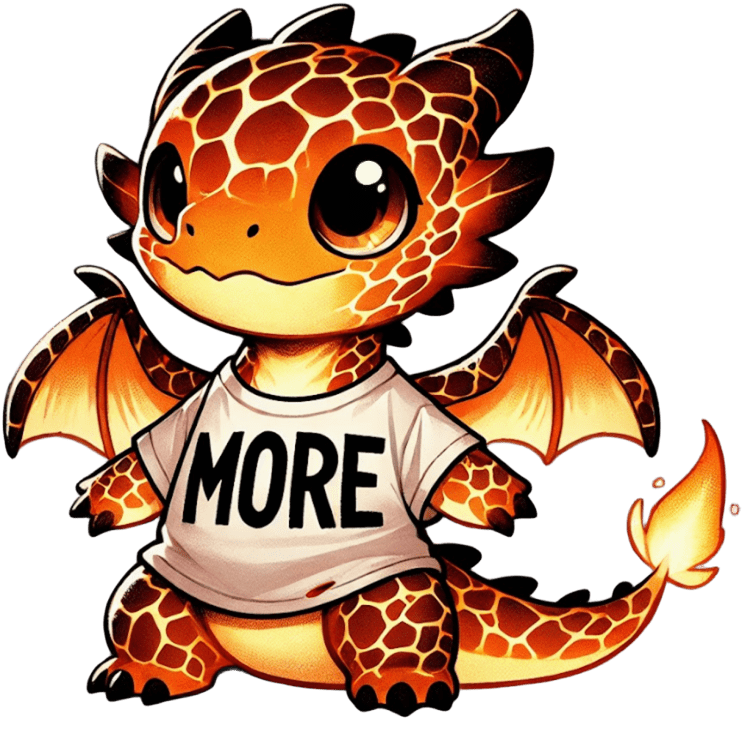}\xspace}

\definecolor{Gray}{gray}{0.93}
\definecolor{ourline}{rgb}{1.0, 0.8, 0.6} 

\newcommand{\tit}[1]{\smallbreak\noindent\textbf{#1.}}
\newcommand{\tinytit}[1]{\noindent\textbf{#1.}}
\newcommand{\ours}{LLaVA-MORE\xspace}

\def \ie {\emph{i.e.}}

\newcommand{\cmark}{\ding{51}}%
\newcommand{\xmark}{\ding{55}}%

\newtcolorbox{idea_1}{colback=ourline!20, colframe=ourline!95!red, 
    title=\faLightbulb\ \, Key Takeaway 1}

\newtcolorbox{idea_2}{colback=ourline!20, colframe=ourline!95!red, 
    title=\faLightbulb\ \, Key Takeaway 2}

\newtcolorbox{idea_3}{colback=ourline!20, colframe=ourline!95!red, 
    title=\faLightbulb\ \, Key Takeaway 3}

\newtcolorbox{idea_4}{colback=ourline!20, colframe=ourline!95!red, 
    title=\faLightbulb\ \, Key Takeaway 4}

\newtcolorbox{idea_5}{colback=ourline!20, colframe=ourline!95!red, 
    title=\faLightbulb\ \, Key Takeaway 5}

\newtcolorbox{idea_6}{colback=ourline!20, colframe=ourline!95!red, 
    title=\faLightbulb\ \, Key Takeaway 6}

\definecolor{Gray}{gray}{0.93}
\newtcolorbox{problems}{colback=white!20, colframe=Gray!75!black, 
    title=\faLightbulb\ \, Open Problems}

\definecolor{iccvblue}{rgb}{0.21,0.49,0.74}
\usepackage[pagebackref,breaklinks,colorlinks,allcolors=iccvblue]{hyperref}

\def\paperID{26} 
\def\confName{ICCV}
\def\confYear{2025}

\title{\titledmodelname: A Comparative Study of LLMs and Visual Backbones for Enhanced Visual Instruction Tuning}

\author{Federico Cocchi\textsuperscript{1,2}\footnotemark[1]\thanks{Equal contribution.}\quad {Nicholas Moratelli}\textsuperscript{1}\footnotemark[1]\quad
{Davide Caffagni}\textsuperscript{1}\footnotemark[1]\quad
{Sara Sarto\textsuperscript{1}\footnotemark[1]}\quad\\  {Lorenzo Baraldi\textsuperscript{1}} \quad {Marcella Cornia\textsuperscript{1}} \quad {Rita Cucchiara\textsuperscript{1,3}} \\
\textsuperscript{1}University of Modena and Reggio Emilia, Italy  \quad
\textsuperscript{2}University of Pisa, Italy \quad
\textsuperscript{3}IIT-CNR, Italy \\
\textsuperscript{1}\small{\texttt{\{name.surname\}@unimore.it}} \quad \textsuperscript{2}\small{\texttt{\{name.surname\}@phd.unipi.it}}
}

\begin{document}
\maketitle
\begin{abstract}
Recent progress in Multimodal Large Language Models (MLLMs) has highlighted the critical roles of both the visual backbone and the underlying language model. While prior work has primarily focused on scaling these components to billions of parameters, the trade-offs between model size, architecture, and performance remain underexplored. Additionally, inconsistencies in training data and evaluation protocols have hindered direct comparisons, making it difficult to derive optimal design choices. In this paper, we introduce \ours, a new family of MLLMs that integrates recent language models with diverse visual backbones. To ensure fair comparisons, we employ a unified training protocol applied consistently across all architectures. Our analysis systematically explores both small- and medium-scale LLMs -- including Phi-4, LLaMA-3.1, and Gemma-2 -- to evaluate multimodal reasoning, generation, and instruction following, while examining the relationship between model size and performance. Beyond evaluating the LLM impact on final results, we conduct a comprehensive study of various visual encoders, ranging from CLIP-based architectures to alternatives such as DINOv2, SigLIP, and SigLIP2. Additional experiments investigate the effects of increased image resolution and variations in pre-training datasets. Overall, our results provide insights into the design of more effective MLLMs, offering a reproducible evaluation framework that facilitates direct comparisons and can guide future model development.  Our source code and trained models are publicly available at: 
{\small{\url{https://github.com/aimagelab/LLaVA-MORE}}}.
\end{abstract}    
\section{Introduction}

\begin{figure}[t]
\centering
\includegraphics[width=\linewidth]{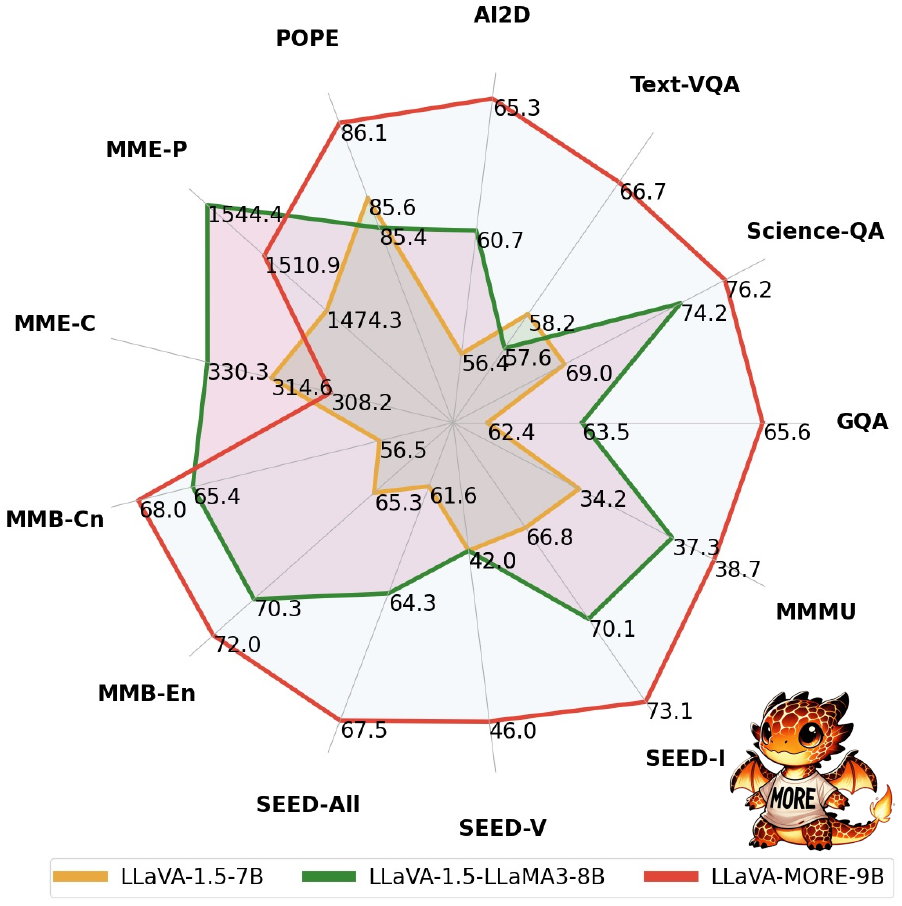}
\vspace{-0.5cm}
\caption{
Performance comparison of the best version of \ours with other LLaVA variants across different benchmarks for multimodal reasoning and visual question answering. 
}
\label{fig:plot}
\vspace{-0.4cm}
\end{figure}

The emergence of Large Language Models (LLMs) with remarkable expressive capabilities has revolutionized the way diverse language-related tasks are approached~\cite{vicuna2023,team2024gemma,touvron2023llama,abdin2024phi}. This advancement has inspired the Computer Vision and Multimedia communities to move beyond traditional text-only paradigms and adopt multiple modalities, including vision, audio, and beyond. Consequently, this shift has led to the emergence of Multimodal Large Language Models (MLLMs)~\cite{caffagni2024r}, which establish sophisticated relationships between concepts across different embedding spaces, enabling richer multimodal understanding. 

Current MLLMs~\cite{alayrac2022flamingo,liu2023visual,liu2024improved,agrawal2024pixtral,chen2024internvl,jiang2024mantis} typically integrate a language model with a visual backbone using specialized adapters that bridge the gap between modalities. While these systems demonstrate impressive performance, the field has converged around a somewhat narrow technical approach, with most implementations leveraging LLaMA-derived LLMs and LLaVA-based training protocols. Additionally, visual encoders based on contrastive training such as CLIP~\cite{radford2021learning} and its derivatives~\cite{tschannen2025siglip,fang2023eva,zhai2023sigmoid} have become the default choice for extracting visual features. These encoders are specifically trained to generate embeddings that seamlessly integrate with language models, further driving their widespread adoption. While contrastive learning has been highly effective in aligning images and text within a shared space, other vision models~\cite{caron2021emerging,oquab2024dinov2} learn robust visual features in a purely self-supervised scheme, without relying on weak supervision from text. These models showcase intriguing emerging properties, and yet their application to MLLMs is relatively understudied.

To address this, our work conducts a comprehensive empirical study that systematically pairs diverse LLMs --ranging from efficient models~\cite{abdin2024phi} to significantly larger architectures~\cite{touvron2023llama,team2024gemma} -- with various visual backbones~\cite{radford2021learning,zhai2023sigmoid,oquab2024dinov2,tschannen2025siglip}.
By exploring different architectural combinations, we aim to uncover the strengths and limitations of various vision-language integration strategies, shedding light on overlooked design choices and their impact on multimodal learning. Fig.~\ref{fig:plot} illustrates a comparison of our best-performing model (\ie, \ours-9B, based on SigLIP2 as visual encoder and Gemma-2-9B as LLM) against LLaVA-based competitors (\ie, LLaVA-1.5-7B~\cite{liu2024improved} and LLaVA-1.5-LLaMA3-8B~\cite{hanoona2024LLaVA}, both using a CLIP-based visual encoder and Vicuna-7B and LLaMA3-8B as LLM, respectively).

To ensure experimental consistency, we follow the established LLaVA~\cite{liu2024improved} methodology, pre-training models on natural language description tasks before applying visual instruction fine-tuning to improve cross-domain generalization and human alignment.
However, recent works not only introduce architectural enhancements but also incorporate specially curated datasets~\cite{deitke2024molmo,bai2023qwen,liu2024nvila}, making fair comparisons across models challenging. To isolate the role of data, in this work we compare LLaVA models pre-trained on different datasets against the same evaluation protocol.

To further explore the impact of training data, we conduct an additional study examining how different types of pre-training data influence multimodal alignment, reasoning ability, and generalization. Specifically, we compare models trained on web-scale image-text corpora, such as LAION~\cite{schuhmann2022laion}, against those leveraging task-specific datasets with richer structural supervision~\cite{liu2024improved}. Additionally, we explore the impact of using Recap-DataComp-1B~\cite{li2024if}, a recaptioned variant designed to enhance image-text alignment. Our results demonstrate that dataset selection significantly affects recognition capabilities and cross-domain transfer performance.

To summarize, this work provides key insights into cross-modal representation learning and offers a practical guide for developing more efficient and effective MLLMs, while challenging conventional assumptions about visual and textual backbones and dataset requirements for optimizing pre-training strategies.

\section{Related Works}

\tinytit{Multimodal Large Language Models}
The rapid advancements in language models, exemplified by GPT-4~\cite{achiam2023gpt} and Gemini~\cite{team2023gemini}, have transformed AI research, particularly through alignment techniques such as instruction tuning~\cite{ouyang2022training,taori2023stanford} and reinforcement learning from human feedback~\cite{stiennon2020learning}. Open-source LLMs~\cite{chung2022scaling,vicuna2023,touvron2023llama,jiang2023identifying,bai2023qwen} as well as more scale-efficient models~\cite{allal2025smollm2,team2024gemma,abdin2024phi,abouelenin2025phi} have significantly driven innovation within the research community. 

Building on these advancements,  MLLMs extend traditional LLMs by incorporating visual perception capabilities, enabling models to process not only textual data but also interpret and reason about visual content. This integration significantly enhances their ability to tackle complex multimodal tasks~\cite{caffagni2024r}. A significant breakthrough in multimodal learning has come from the LLaVA models family~\cite{liu2023visual,liu2024improved}, which introduced visual instruction tuning to MLLMs. By leveraging a GPT-4-curated text-only dataset, LLaVA models effectively aligned visual and textual representations, leading to substantial improvements in multimodal performance. This approach has since become a foundational paradigm for training multimodal models, driving further advancements in the field.

Recent works have not only refined standard multimodal capabilities but also expanded their focus to a broader range of vision-centric benchmarks and diverse settings~\cite{bai2023qwen}. To achieve state-of-the-art results, some approaches prioritize enhancing the visual encoding phase, eventually fine-tuning~\cite{mckinzie2024mm1} a vision encoder or even training it from scratch~\cite{agrawal2024pixtral}. Others focus instead on scaling up spatial and temporal resolutions or improving visual token compression~\cite{liu2024nvila, bolya2022token} for better efficiency.

Despite the impressive performance of these models, a significant drawback lies in their reliance on specially curated datasets during training~\cite{laurenccon2023obelics,deitke2024molmo,bai2023qwen,liu2024nvila,laurenccon2024matters}, which complicates fair evaluation and direct comparisons across different architectures and settings.

\tit{Large Language Models and Recent Advances}
Despite the development of various LLM architectures~\cite{zhang2022opt,wang2023magneto,raffel2020exploring,chung2022scaling,xue2020mt5}, the LLaMA family~\cite{touvron2023llama,touvron2023llama2,grattafiori2024llama} remains a popular choice thanks to its open-access nature, reliance on publicly available model weights, and availability in multiple sizes. Variants such as Alpaca~\cite{taori2023stanford} and Vicuna~\cite{vicuna2023} further build upon LLaMA, enhancing its instruction-following and conversational abilities. 

Recent research has focused on developing lightweight LLMs that strike a balance between efficiency and performance~\cite{allal2025smollm2}. A notable example is the Gemma family~\cite{team2024gemma1,team2024gemma}, which builds upon Gemini~\cite{team2023gemini} and is designed to provide strong reasoning and comprehension skills at various computational scales.
Another significant contribution comes from the DeepSeek models~\cite{liu2024deepseek,guo2025deepseek}, which are characterized by their efficient training using reinforcement learning and their optimized inference processes.

A major breakthrough in LLM development is the introduction of LLaMA-3~\cite{grattafiori2024llama}, offering 8B to 405B parameter models with enhanced multilingual, coding, and reasoning abilities. Its improvements stem from rigorous data curation, pre-processing, and filtering, highlighting the critical role of high-quality training data in boosting efficiency and accuracy. This principle is further exemplified by the Phi models~\cite{abouelenin2025phi,textbooks2,abdin2024phi}, which outperforms larger LLMs across multiple benchmarks. Among these, Phi-4~\cite{abdin2024phi} pushes the boundaries of small-scale models through synthetic data generation, optimized training, and advanced post-training techniques, demonstrating that strategic data refinement can rival sheer model scaling.

\tit{Visual Backbones}
Beyond advancements in language modeling, the visual backbone plays a crucial role in the performance of MLLMs by extracting and encoding image features for multimodal reasoning and understanding. The most widely used visual encoders are Vision Transformers (ViTs) trained with CLIP-style contrastive learning, with CLIP ViT-L~\cite{radford2021learning} being the most commonly adopted architecture.
Building on the success of CLIP, several studies have explored similar methodologies that leverage the inherent alignment of cross-modal embeddings. 

The SigLIP family~\cite{zhai2023sigmoid,tschannen2025siglip} enhances CLIP by refining the training objective and filtering high-quality data, improving vision-language performance. Likewise, DINO models~\cite{caron2021emerging, oquab2024dinov2, darcetvision} introduces an automated pipeline for dataset filtering and rebalancing, leveraging vast collections of uncurated images to produce a diverse set of pre-trained visual models.
Overall, stronger visual encoders generally boost MLLMs~\cite{li2023blip}. Expanding on this,
some recent approaches~\cite{lin2023sphinx, gao2024sphinxx,lu2024deepseek,tong2024cambrian,tong2024eyes} investigate multi-backbone fusion strategies, combining different encoders to enhance feature diversity and robustness. Other solutions, such as PaLI~\cite{chen2023pali,chen2023palix}, tackle the imbalance between visual and language parameters by scaling the vision backbone to billions of parameters.
However, recent works~\cite{shi2024we} suggest that multi-scale feature extraction with smaller ViT-based models can outperform larger encoders, offering a more efficient alternative to scaling up visual encoders.

\begin{figure}[t]
    \centering
    \includegraphics[width=0.99\linewidth]{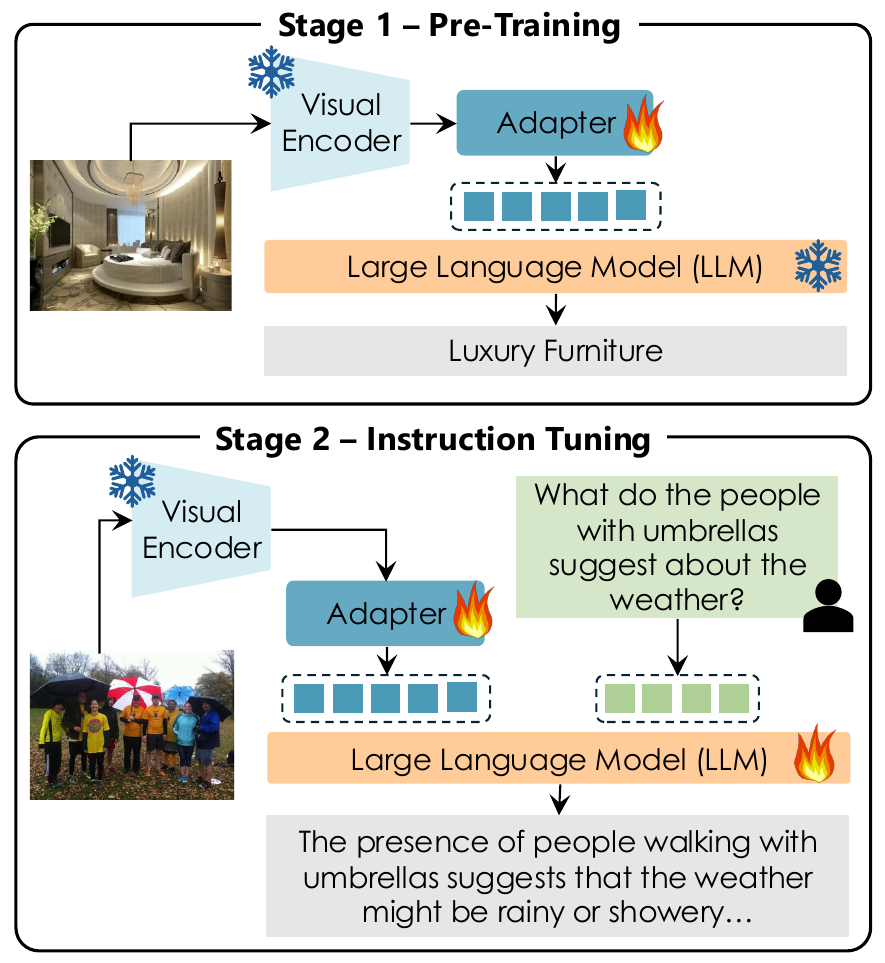}
    \vspace{-0.2cm}
    \caption{Overview of the \ours architecture. Our approach follows the standard LLaVA framework with a two-stage training process. The first stage aligns the visual features to the underlying LLM, ensuring effective cross-modal representation. The second stage enhances the MLLM conversational capabilities through visual instruction tuning. Following this paradigm, we systematically compare different LLM and visual encoder choices to evaluate their impact on various multimodal tasks.}
    \label{fig:model}
    \vspace{-0.4cm}
\end{figure}

\section{Overall Architecture}
Following the LLaVA architecture, a typical MLLM consists of three fundamental components: a large language model backbone for user interaction and generating text, one or more visual encoders to process visual input and extract features, and at least one vision-to-language adapter to bridge the gap between visual and textual modalities~\cite{caffagni2024r}.
The visual encoder provides essential global visual features to the LLM, with CLIP-based architectures~\cite{radford2021learning, wortsman2022robust} being the most commonly used. 
In this setup, the visual encoder is typically pre-trained on another task and kept frozen for the entire training of the MLLM.

In this paper, we propose \ours, a new family of models that extend the standard LLaVA architecture by combining the visual encoder with various LLMs, ranging from small- to medium-scale models. As small-scale models, we utilize Gemma-2 2B~\cite{team2024gemma} and Phi-4-Mini~\cite{abdin2024phi} (with 3.8B parameters), both designed for strong reasoning scalability, effectively challenging larger models. For medium-scale models, we select the variant of Gemma-2~\cite{team2024gemma} with 9B parameters alongside two recent architectures: the original LLaMA-3.1~\cite{grattafiori2024llama} LLM with 8B parameters and its distilled DeepSeek-R1 version~\cite{guo2025deepseek} (\ie, DeepSeek-R1-Distill-LLaMA-8B). For each LLM category, we assess the impact of varying the visual backbone, with the aim to identify the optimal configuration. Specifically, our study compares the standard CLIP ViT-L/14 encoder employed in the LLaVA-1.5 model~\cite{liu2024improved} with two variants of DINOv2 differentiated by the presence or the absence of visual register tokens~\cite{oquab2024dinov2, darcetvision}, known for their strong, semantically rich visual features, as well as SigLIP~\cite{zhai2023sigmoid} and its more advanced successor, SigLIP2~\cite{tschannen2025siglip}.

To train our models, we follow the two-stage training paradigm commonly used in the literature. However, our approach stands out by applying a consistent training and evaluation strategy across all models, ensuring fairness in comparisons. In the first stage, only the vision-to-language adapter is optimized to align the image features with the text embedding space. In the second stage, visual instruction-following training is conducted to enhance multimodal conversational capabilities. During this phase, the parameters of both the multimodal adapter and the LLM are updated. An overview of the overall architecture and the two-stage training process is shown in Fig.~\ref{fig:model}.

\section{Experimental Evaluation}

\subsection{Implementation Details}
\tinytit{Architectural Components}
The \ours family follows LLaVA architecture, employing CLIP ViT-L/14@336 as the visual backbone while varying the underlying language model. We categorize the selected LLMs into two groups based on scale: small-scale models, including Gemma-2-2B and Phi-4-3.8B, and medium-scale models, such as LLaMA-3.1-8B, DeepSeek-R1-Distill-LLaMA-8B, and Gemma-2-9B. To further investigate the impact of the visual backbone, we conduct experiments on the best-performing models, replacing CLIP~\cite{radford2021learning} with alternative vision encoders, including DINOv2~\cite{oquab2024dinov2}, SigLIP~\cite{zhai2023sigmoid}, and SigLIP2~\cite{tschannen2025siglip}. Additionally, we examine the effects of applying Scaling on Scales (S$^2$)~\cite{shi2024we} to both the CLIP and SigLIP2 architectures. Finally, motivated by the LLaVA-1.5 framework and insights from~\cite{chen2020simple, chen2020improved}, which highlight the advantages of using MLPs over linear projections in self-supervised learning, we adopt a two-layer MLP as the vision-language adapter to enhance multimodal fusion.

\tit{Training Details}
Considering the LLaVA framework, we adopt a two-stage training strategy. In the first stage, only the weights of the adapter are updated using image-caption pairs as training data. Specifically, the caption style follows the alt-text structure as in web-scale multimodal datasets. The models are trained for one full epoch, covering a total of 558k samples from a combination of different sources (\ie, LAION~\cite{schuhmann2022laion}, CC3M~\cite{changpinyo2021conceptual}, and SBU~\cite{ordonez2011im2text}).
In the second step, we fine-tune the model on high-quality visual instruction-following data to improve its multimodal reasoning capabilities. This sequential training approach has been shown to significantly improve performance on downstream tasks~\cite{liu2023visual}.
Notably, the next token prediction is used as the loss function in both training phases. 
\ours models are trained with the same set of hyperparameters as LLaVA-1.5 to ensure consistency and comparability. In particular, we employ a global batch size of 256 during pre-training and 128 during the visual instruction tuning phase. All experiments are run in a multi-GPU, multi-node configuration with a total of 16 A100 64GB NVIDIA GPUs.

\begin{table*}[t] 
\small
\centering
\setlength{\tabcolsep}{.25em}
\resizebox{\linewidth}{!}{
\begin{tabular}{lc cccc cc ccccccccc}
\toprule
& & \multicolumn{4}{c}{\textbf{VQA Benchmarks}} & & & \multicolumn{9}{c}{\textbf{MLLM Benchmarks}} \\
\cmidrule{3-6} \cmidrule{8-17}
& & GQA & Science-QA & TextVQA & AI2D & & & POPE & MME-P & MME-C & MMB-Cn & MMB-En & SEED-All & SEED-V & SEED-I & MMMU \\
\midrule
\rowcolor{Gray}
\multicolumn{17}{l}{\textit{Small-Scale MLLMs}} \\
LLaVA-Phi-2.7B~\cite{zhu2024llava} && - & 68.4 & - & - &&& 85.0 & 1335.1 & - & - &	59.8 & - & - & - & - \\
\midrule
\rowcolor{ourline}
\multicolumn{17}{l}{\textbf{\ours (Ours)}} \\
\hspace{0.3cm}Gemma-2-2B~\cite{team2024gemma} && \textbf{62.4} & 71.1 & \textbf{54.4} & 57.1 &&& \textbf{86.0} & \textbf{1401.1} & \textbf{337.8} & \textbf{65.8} & 53.3 & 62.2 & 41.9 & 67.6  & 33.4 \\
\hspace{0.3cm}Phi-4-3.8B~\cite{abdin2024phi} && 62.1 & \textbf{71.3} & 54.0 & \textbf{61.1} &&& 85.9 & 1372.2 & 281.1 & 64.2 & \textbf{69.2} & \textbf{63.5} & \textbf{42.3} & \textbf{69.1} & \textbf{38.8} \\
\midrule
\midrule
\rowcolor{Gray}
\multicolumn{17}{l}{\textit{Medium-Scale MLLMs}} \\
LLaVA-1.5-7B~\cite{liu2024improved}  && 62.4 & 69.0 & 58.2 & 56.4 &&& 85.6 & 1474.3 & 314.6 & 56.5 & 65.3 & 
61.6 & 42.0 & 66.8 & 34.2\\
LLaVA-1.5-LLaMA3-8B~\cite{hanoona2024LLaVA} && 63.5 & 74.2 & 57.6 & 60.7 &&&  85.4 & \textbf{1544.4} & 330.3 & 65.4 & 70.3 & 64.3 & 42.0 & \textbf{70.1} & 37.3 \\
\midrule
\rowcolor{ourline}
\multicolumn{17}{l}{\textbf{\ours (Ours)}} \\
\hspace{0.3cm}LLaMA-3.1-8B~\cite{grattafiori2024llama} && 63.6 & \textbf{76.3} & 58.4 & 61.8 &&&  85.1 & 1531.5 & \textbf{353.3} & \textbf{68.2} & \textbf{72.4} & 64.1 & 42.4 & 69.8 & \textbf{39.4} \\
\hspace{0.3cm}DeepSeek-R1-Distill-LLaMA-8B~\cite{guo2025deepseek} &&  63.0 & 74.5 & 56.3 & 58.8 &&&  85.1 & 1495.1 & 295.0 & 66.8 & 61.3 & 63.5  & 43.5 & 68.6  & 38.1 \\
\hspace{0.3cm}Gemma-2-9B~\cite{team2024gemma} && \textbf{64.2} & 75.4 & \textbf{60.7} & \textbf{64.8} &&& \textbf{86.8} & 1522.5 & 307.5 & 65.9 & {71.9} & \textbf{64.5} & \textbf{44.1} & \textbf{69.9} & 37.9 \\
\bottomrule
\end{tabular}
}
\vspace{-0.1cm}
\caption{Performance analysis when changing the underlying LLMs. Results are reported considering both small- and medium-scale LLMs, comparing \ours with existing LLaVA-based variants. All models employ the CLIP ViT-L/14@336 as the visual backbone.}
\label{tab:diff_llms}
\vspace{-0.3cm}
\end{table*}

\subsection{Evaluation Benchmarks}
We evaluate the \ours family on a diverse set of task-oriented and instruction-following benchmarks.

\tit{VQA Benchmarks} These primarily assess the model ability to answer questions based on visual inputs. 
For the VQA setting, we consider the following datasets:
\begin{itemize}[leftmargin=*]
\item \textbf{GQA}~\cite{hudson2019gqa} is based on Visual Genome scene graph annotations~\cite{krishna2017visual} and comprises 113k images and 22M questions focusing on scene understanding and compositionality. Results are reported on the test split, which represents 10\% of the total image set.
\item \textbf{ScienceQA}~\cite{lu2022learn} evaluates models with challenging multimodal multiple-choice questions across three diverse domain subjects (\ie, natural science, language science, and social science), 26 topics, 127 categories, and 379 skills. Each question is annotated with explanations linked to relevant lectures from elementary and high school science curricula. We report results on the test set which includes 4,241 examples.

\item \textbf{TextVQA}~\cite{singh2019towards} is a dataset built on Open Images~\cite{kuznetsova2020open} designed to evaluate the OCR capabilities of vision-and-language models. In our experiments, we employ the validation set which comprises 5,734 samples.
\item \textbf{AI2D}~\cite{kembhavi2016diagram} is a comprehensive collection of diagrams specifically designed for educational and research purposes. It consists of over 5,000 grade school science diagrams, which cover a wide range of topics and are accompanied by more than 15,000 diverse and richly formulated multiple-choice questions and answers.
 
\end{itemize}

\tit{MLLM Benchmarks}
These evaluate broader multimodal language understanding and reasoning capabilities. Specifically, we consider the following benchmarks:
\begin{itemize}[leftmargin=*]
    \item \textbf{POPE}~\cite{li2023evaluating} is a benchmark for evaluating object hallucinations in MLLM generations. It includes several subsets, namely random, popular, and adversarial, which are generated using various sampling methodologies. The dataset consists of 8,910 binary classification queries, enabling thorough analysis of the object hallucination phenomena in MLLMs.
    \item \textbf{MME}~\cite{fu2023mme} is designed to measure proficiency in various communication modalities through 14 diverse tasks. These tasks assess comprehension and manipulation in areas such as quantification, spatial reasoning, and color identification. Overall, it contains 2,374 samples.
    \item \textbf{MMBench (MMB)}~\cite{liu2023mmbench} includes approximately 3,000 multiple-choice questions, distributed across a collection of 20 distinct domains and understanding capability. Questions are designed to assess the effectiveness of MLLMs across various task paradigms. These capabilities are systematically structured into a hierarchical taxonomy, encompassing broad categories like perception and reasoning, as well as finer-grained skills such as object localization and attribute inference.
    \item \textbf{SEED-Bench (SEED)}~\cite{li2023seed} evaluates MLLMs across 12 dimensions, including scene understanding, OCR, and action recognition. The dataset comprises 19k multiple-choice questions curated by human annotators.
    \item \textbf{MMMU}~\cite{yue2023mmmu} is a challenging benchmark for multimodal models, focusing on massive multi-discipline tasks demanding college-level subject knowledge. It consists of 900 validation samples drawn from university textbooks or online courses spanning six main disciplines. Questions may include multiple images interleaved with text. Evaluation includes exact and word matching for multiple-choice and open-ended questions, and models are tested in zero or few-shot settings.
\end{itemize}

\vspace{-0.2cm}
\subsection{Assessing the Optimal LLM Choice}
\ours extends the widely recognized LLaVA architecture by incorporating small- and medium-scale LLMs. As shown in Table~\ref{tab:diff_llms}, our experiments first investigate the relationship between model size and performance across various benchmarks.
Among the small-scale LLMs, we compare the LLaVA version based on Phi-2.7B~\cite{zhu2024llava} with our versions based on Phi-4-3.8B and Gemma-2-2B. Notably, both versions of \ours consistently outperform the existing baseline across multiple benchmarks. Notably, Phi-4-3.8B achieves the highest scores on most benchmarks, particularly excelling in the MMMU performance, where it surpasses Gemma-2-2B by 5.4\%. Similar improvements are also evident in the SEED dataset, where Phi4-3.8B achieved significantly higher performance than other small-scale LLMs. These results underscore Phi-4-3.8B’s superior reasoning and generalization capabilities.

Conversely, among medium-scale models, LLaVA-1.5-7B and LLaVA-1.5-LLaMA3-8B provide stronger baselines, with LLaVA-1.5-LLaMA3-8B achieving the highest score in MME-P (1544.4). However, our \ours models consistently surpass these baselines, demonstrating superior performance across both VQA and MLLM benchmarks. In particular, Gemma-2-9B emerges as the best-performing model, especially excelling in VQA benchmarks. It achieves the highest scores on GQA and AI2D, significantly outperforming both baselines and other models within the \ours family. In MLLM benchmarks, \ours combined with LLaMA-3.1-8B demonstrates instead strong capabilities on the MMB dataset, showing good results in multiple-choice question settings.
Notably, our models exhibit less sensitivity to object hallucinations, as seen by the results achieved on the POPE benchmark, and demonstrate superior performance on the SEED dataset, further highlighting their robustness in multimodal reasoning tasks.

Comparing small- and medium-scale models, it is noteworthy that some small-scale \ours models outperform even medium-scale baselines like LLaVA-1.5-7B. For instance, in Science-QA and AI2D, \ours with Phi-4-3.8B surpasses both the baseline and \ours with DeepSeek-R1-Distill-LLaMA-8B.

This trend extends to MLLM benchmarks, where \ours models demonstrate competitive performance in MMB and SEED, further highlighting their efficiency and strong reasoning capabilities despite their smaller size.
Overall, our results emphasize that scaling up is not the only path to better performance, as architectural choices and fine-tuning strategies significantly impact model effectiveness across different benchmarks.

\begin{problems}
\centering
Multimodal reasoning needs improvement through training strategies specifically designed for cross-modal understanding. 
\end{problems}

\begin{idea_1}
\centering
Recent small-scale models are comparable to medium-scale models of the previous generation.
\end{idea_1}

\begin{table*}[t] 
\vspace{-0.15cm}
\small
\centering
\setlength{\tabcolsep}{.2em}
\resizebox{\linewidth}{!}{
\begin{tabular}{lccc cccc cc ccccccccc}
\toprule
& & & & \multicolumn{4}{c}{\textbf{VQA Benchmarks}} & & & \multicolumn{9}{c}{\textbf{MLLM Benchmarks}} \\
\cmidrule{5-9} \cmidrule{11-19}
& \textbf{Resolution} & \textbf{\# Tokens} & & GQA & Science-QA & TextVQA & AI2D & & & POPE & MME-P & MME-C & MMB-Cn & MMB-En & SEED-All & SEED-V & SEED-I & MMMU \\
\midrule
\rowcolor{ourline}
\multicolumn{19}{l}{\textbf{\ours-3.8B (Ours)}} \\
\hspace{0.3cm}CLIP ViT-L/14~\cite{radford2021learning} & 336$^2$ & 576 & & 62.1 & 71.3 & 54.0 & 61.1 &&& 85.9 & 1372.2 & 281.1 & 64.2 & 69.2 & 63.5 & 42.3 & 69.1 & 38.8 \\
\hspace{0.3cm}DINOv2 ViT-L/14~\cite{oquab2024dinov2} & 224$^2$ & 256 & & 60.9 & 66.6 & 41.4 & 58.2 &&& 85.5 & 1236.6 & 281.1 & 53.8 & 58.9 & 59.8 & 40.6 & 64.8 & 37.9 \\
\hspace{0.3cm}DINOv2$_\text{reg}$ ViT-L/14~\cite{darcetvision} & 224$^2$ & 256 & & 60.4 & 69.0 & 41.3 & 56.4 &&& 85.2 & 1263.2 & \textbf{288.2} & 57.4 &	51.4 & 58.7 & 41.4 & 63.2 & 38.6 \\
\hspace{0.3cm}SigLIP ViT-L/14~\cite{zhai2023sigmoid} & 384$^2$ & 729 & & \textbf{63.6} & \textbf{73.8} & 57.6 & \textbf{62.9} &&& 86.4 & 1379.0 & 282.9 & 66.5 & \textbf{71.4} & 65.7 & 46.4 & 70.8 & \textbf{40.0}\\
\hspace{0.3cm}SigLIP2 ViT-L/14~\cite{tschannen2025siglip} & 384$^2$ & 729 & & 63.4 & 71.8 & \textbf{59.7} & \textbf{62.9} &&& \textbf{86.5} & \textbf{1406.7} & 282.5 & \textbf{66.8}	& 69.8 & \textbf{66.4} & \textbf{47.4} & \textbf{71.4} & 38.8 \\
\midrule
\rowcolor{ourline}
\multicolumn{19}{l}{\textbf{\ours-9B (Ours)}} \\
\hspace{0.3cm}CLIP ViT-L/14~\cite{radford2021learning} & 336$^2$ & 576 & & {64.2} & 75.4 & {60.7} & {64.8} &&& \textbf{86.8} & \textbf{1522.5} & 307.5 & 65.9 & {71.9} & {64.5} & {44.1} & {69.9} & 37.9 \\
\hspace{0.3cm}DINOv2 ViT-L/14~\cite{oquab2024dinov2} & 224$^2$ & 256 & & 63.1 & 71.5 & 48.1 & 61.3 &&& 85.3 & 1394.4 & \textbf{334.3} & 56.4 &	63.8 & 61.0 & 40.6 & 66.4 & 38.7 \\
\hspace{0.3cm}DINOv2$_\text{reg}$ ViT-L/14~\cite{darcetvision} & 224$^2$ & 256 & & 62.8 & 69.1 & 47.9 & 59.1	 &&& 84.0 & 1413.9 & 295.4	 & 60.1 & 53.8 & 60.1 & 42.3 & 64.7 & 38.3 \\
\hspace{0.3cm}SigLIP ViT-L/14~\cite{zhai2023sigmoid} & 384$^2$ & 729 & & 64.8 & \textbf{76.3} & 63.9 & 64.7 &&& 86.1 & 1487.9 & 299.3 & \textbf{69.1} & \textbf{74.4} & 66.6 & \textbf{46.6} & 71.9 & \textbf{39.7} \\
\hspace{0.3cm}SigLIP2 ViT-L/14~\cite{tschannen2025siglip} & 384$^2$ & 729 & & \textbf{65.6} & 76.2 & \textbf{66.7} & \textbf{65.3}	&&& 86.1 & 1510.9 & 308.2 & 68.0 & 72.0 & \textbf{67.5} & 46.0 & \textbf{73.1} & 38.7 \\
\bottomrule
\end{tabular}
}
\vspace{-0.1cm}
\caption{Performance analysis with varying visual backbones. Results are reported for the best small- and medium-scale \ours configurations using Phi-4-3.8B and Gemma-2-9B, respectively. Input resolution and the number of visual tokens are also included.}
\vspace{-0.1cm}
\label{tab:diff_backbone}
\end{table*}

\begin{table*}[t] 
\small
\centering
\setlength{\tabcolsep}{.25em}
\resizebox{\linewidth}{!}{
\begin{tabular}{lccccc cccc cc ccccccccc}
\toprule
& & & & & & \multicolumn{4}{c}{\textbf{VQA Benchmarks}} & & & \multicolumn{9}{c}{\textbf{MLLM Benchmarks}} \\
\cmidrule{6-10} \cmidrule{12-21}
& & \textbf{S$^\text{2}$} &  \textbf{Resolution} & \textbf{\# Tokens} & & GQA & Science-QA & TextVQA & AI2D & & & POPE & MME-P & MME-C & MMB-Cn & MMB-En & SEED-All & SEED-V & SEED-I & MMMU \\
\midrule
\rowcolor{ourline}
\multicolumn{21}{l}{\textbf{\ours-3.8B (Ours)}} \\
\hspace{0.3cm}CLIP ViT-L/14~\cite{radford2021learning} & & \xmark & 336$^2$ & 576 & & 62.1 & 71.3 & 54.0 & 61.1 &&& 85.9 & \textbf{1372.2} & \textbf{281.1} & 64.2 & \textbf{69.2} & 63.5 & 42.3 & 69.1 & 38.8 \\
\hspace{0.3cm}CLIP ViT-L/14~\cite{shi2024we} & & \cmark & 336$^2\times$14 & 576 && \textbf{62.7} & \textbf{71.8} & \textbf{54.9} & \textbf{61.7} &&& \textbf{86.9} & 1366.8 & 263.9 & \textbf{68.6} & 64.4 & \textbf{64.2} & \textbf{43.1} & \textbf{69.8} & \textbf{39.8} \\
\midrule
\hspace{0.3cm}SigLIP2 ViT-L/14~\cite{tschannen2025siglip} & & \xmark & 384$^2$ & 729 & & 63.4 & 71.8 & 59.7 & 62.9 &&& 86.5 & 1406.7 & 282.5 & 66.8 & 69.8 & 66.4 & 47.4 & 71.4 & \textbf{38.8} \\
\hspace{0.3cm}SigLIP2 ViT-L/14~\cite{shi2024we} & & \cmark & 384$^2\times$14 & 729 & & \textbf{64.1} & \textbf{72.2} & \textbf{61.5} & \textbf{63.2} &&& \textbf{87.4} & \textbf{1466.7} & \textbf{331.1} & \textbf{69.2} & \textbf{70.7} & \textbf{67.0} & \textbf{47.6} & \textbf{72.1} & 38.7 \\
\midrule
\rowcolor{ourline}
\multicolumn{21}{l}{\textbf{\ours-9B (Ours)}} \\
\hspace{0.3cm}CLIP ViT-L/14~\cite{radford2021learning} & & \xmark & 336$^2$ & 576 & & 64.2 & \textbf{75.4} & 60.7 & \textbf{64.8} &&& \textbf{86.8} & \textbf{1522.5} & 307.5 & 65.9 & \textbf{71.9} & 64.5 & 44.1 & 69.9 & 37.9 \\
\hspace{0.3cm}CLIP ViT-L/14~\cite{shi2024we} & & \cmark & 336$^2\times$14 & 576 & & \textbf{65.2} & 73.7 & \textbf{63.4} & 64.2 &&& 86.1 & 1495.9 & \textbf{331.8} & \textbf{68.6} & 70.2 & \textbf{65.1} & \textbf{45.2} & \textbf{70.4} & \textbf{39.0} \\
\midrule
\hspace{0.3cm}SigLIP2 ViT-L/14~\cite{tschannen2025siglip} & & \xmark & 384$^2$ & 729 & & 65.6 & \textbf{76.2} & 66.7 & \textbf{65.3} &&& 86.1 & 1510.9 & 308.2 & 68.0 & 72.0 & \textbf{67.5} & \textbf{46.0} & \textbf{73.1} & 38.7 \\
\hspace{0.3cm}SigLIP2 ViT-L/14~\cite{shi2024we} & & \cmark & 384$^2\times$14 & 729 & & \textbf{65.9} & 74.9 & \textbf{68.1} & 64.1 &&& \textbf{86.7} & \textbf{1557.6} & \textbf{320.7} & \textbf{68.6} & \textbf{73.5} & 67.2 & 45.6 & 72.9 & \textbf{40.2} \\
\bottomrule
\end{tabular}
}
\vspace{-0.1cm}
\caption{
Performance analysis when applying the S$^2$ multi-scale visual processing~\cite{shi2024we}. Results are reported considering the best small- and medium-scale \ours configurations with Phi-4-3.8B and Gemma-2-9B, using both CLIP and SigLIP2 visual encoders.
}
\label{tab:diff_backbone_s2}
\vspace{-0.1cm}
\end{table*}

\subsection{Changing the Visual Backbone}
We then investigate which visual backbone is more promising for building MLLMs. To this end, we select the best small- and medium-scale models resulting from Table~\ref{tab:diff_llms} and study how their performance is affected when the visual encoder is varied. In detail, we opt for Phi-4-3.8B as the small-scale LLM (\ie, \ours-3.8B), and Gemma-2-9B as the medium-scale one (\ie, \ours-9B). As per the visual backbone, in Table~\ref{tab:diff_backbone}, we include four pre-trained ViT-based models, in addition to the standard CLIP used by LLaVA. All models share the ViT-L/14 architecture, and yet there are striking differences in terms of 
training data, input image resolutions, and on the pre-training strategies. 
In particular, we can delineate two pre-training paradigms: DINOv2~\cite{oquab2024dinov2}, eventually enhanced with registers~\cite{darcetvision}, has been pre-trained with self-supervision and knowledge distillation on 142M images, while the other visual encoders leverage the weak supervision of noisy image-text pairs during pre-training. Specifically, CLIP~\cite{radford2021learning} learns robust visual models via contrastive learning, while SigLIP~\cite{zhai2023sigmoid} exploits the sigmoid loss to improve cross-modal alignment. The recent SigLIP2~\cite{tschannen2025siglip} builds upon SigLIP by incorporating additional training objectives, including image captioning, self-distillation, and image-masked prediction.

From Table~\ref{tab:diff_backbone}, we observe that image-text pre-trained visual backbones consistently outperform DINOv2 backbones at both small and medium scales. Moreover, adding register tokens in DINOv2 does not help in reducing the gap. We retain that, thanks to their pre-training, CLIP and SigLIP provide visual features that are readily aligned with text, simplifying the role of the multimodal adapter of making them understandable by the LLM.

The only exception is MME-C, where DINOv2 with registers scores the highest at the 3.8B scale, and DINOv2 is the best at the 9B scale with 334.3 points. 
Among the models that benefited from image-text pre-training, SigLIP shows a substantial improvement over CLIP at both scales, highlighting its effectiveness despite introducing a computational overhead—specifically, it forces the LLM to process approximately 26\% more visual tokens due to the use of higher-resolution inputs.
Despite its more complicated training recipe, the new SigLIP2 performs on par with the original SigLIP at the 3.8B scale but records an average 0.4\% gain over SigLIP at the 9B scale. For this reason, we will always include SigLIP2 as the visual backbone in all the subsequent experiments. One key finding from our evaluation is that SigLIP-based visual backbones establish a new performance frontier for MLLMs across most benchmarks. Beyond the advantage conferred by increased resolution, we attribute much of this success to the massive billion-scale image-text pre-training enabled by the sigmoid loss of SigLIP, compared to the 400M image-text pairs seen by CLIP during pre-training. 

\begin{idea_2}
\centering
Visual backbones trained with contrastive learning outperform self-supervised visual encoders.
\end{idea_2}

\begin{idea_3}
\centering
Different versions of SigLIP consistently outperform other visual backbones.
\end{idea_3}

\subsection{Changing the Image Resolution}
The choice of a robust visual backbone and the appropriate LLM are critical factors in enhancing model performance. However, the resolution of the input image plays a similarly important role in visual understanding. Higher image resolutions provide more fine-grained visual information, which can significantly aid the MLLM in better interpreting the content and thereby improving performance. 

\begin{table*}[t] 
\vspace{-0.15cm}
\small
\centering
\setlength{\tabcolsep}{.22em}
\resizebox{0.9\linewidth}{!}{
\begin{tabular}{lc cccc cc ccccccccc}
\toprule
& & \multicolumn{4}{c}{\textbf{VQA Benchmarks}} & & & \multicolumn{9}{c}{\textbf{MLLM Benchmarks}} \\
\cmidrule{3-7} \cmidrule{9-17}
& & GQA & Science-QA & TextVQA & AI2D & & & POPE & MME-P & MME-C & MMB-Cn & MMB-En & SEED-All & SEED-V & SEED-I & MMMU \\
\midrule
\rowcolor{ourline}
\multicolumn{17}{l}{\textbf{\ours-3.8B (Ours)}} \\
\hspace{0.3cm}LLaVA Pre-Train LCS (558k) & & 63.4 & 71.8 & 59.7 & 62.9 &&& 86.5 & 1406.7 & 282.5 & 66.8	& 69.8 & 66.4 & 47.4 & 71.4 & 38.8 \\
\hspace{0.3cm}LAION (558k) & & 64.3 & \textbf{72.5} & \textbf{62.3} & \textbf{65.2} &&& \textbf{86.8} & \textbf{1453.2} & 287.1 & \textbf{67.2} & \textbf{72.3} & 66.6 & 46.4 & 71.9 & \textbf{39.7} \\
\hspace{0.3cm}Recap (558k) & & \textbf{64.6} & 71.7 & 61.4 & 64.5 &&& 86.5 & 1428.7 & \textbf{297.9} & 67.1 & 71.6 & 67.3 & \textbf{47.7} & 72.5 & 39.0 \\
\hspace{0.3cm}LAION+Recap (558k) & & \textbf{64.6} & 71.8 & 61.3 & 63.9 &&& 86.6 & 1425.8 & 297.5 & 65.8 & 71.7 & \textbf{67.6} & 47.5 & \textbf{72.9} & 39.2 \\
\midrule
\rowcolor{ourline}
\multicolumn{17}{l}{\textbf{\ours-9B (Ours)}} \\
\hspace{0.3cm}LLaVA Pre-Train LCS (558k) & & 65.6 & 76.2 & 66.7 & \textbf{65.3}	&&& 86.1 & 1510.9 & 308.2 & 68.0 & 72.0 & 67.5 & 46.0 & 73.1 & 38.7 \\
\hspace{0.3cm}LAION (558k) & & 65.6 & 76.0 & 67.0 & 65.1 &&& \textbf{86.9} & \textbf{1579.8} & \textbf{350.7} & 68.5 & 73.9 & 67.4 & 45.4 & \textbf{73.2} & 41.1 \\
\hspace{0.3cm}Recap (558k) & & \textbf{65.9} & 76.2 & 67.2 & 64.3 &&& \textbf{86.9} & 1540.8 & 318.2 & \textbf{69.4} & \textbf{74.7} & \textbf{67.6} & \textbf{47.2} & 73.0 & 40.2 \\
\hspace{0.3cm}LAION+Recap (558k) & & 65.8 & \textbf{77.1} & \textbf{67.8} & 65.2 &&& 86.7 & 1537.8 & 335.4 & 65.9 & 73.5 & 67.4 & 45.2 & 73.1 & \textbf{41.2} \\
\bottomrule
\end{tabular}
}
\vspace{-0.1cm}
\caption{
Performance analysis when changing the training dataset during the first pre-training stage. Results are reported for the best small- and medium-scale \ours configurations using Phi-4-3.8B and Gemma-2-9B.}
\vspace{-0.2cm}
\label{tab:diff_dataset}
\end{table*}

\begin{figure*}[t]
    \centering
    \includegraphics[width=\linewidth]{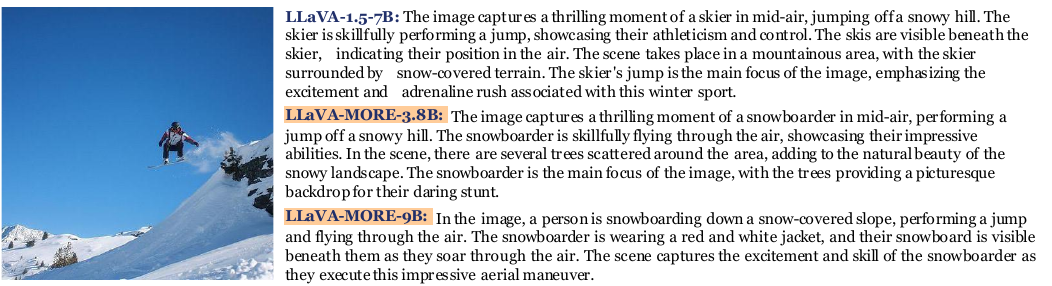}
    \vspace{-0.5cm}
    \caption{Qualitative of image descriptions generated by three MLLMs (LLaVA-1.5-7B, LLaVA-MORE-3.8B, and LLaVA-MORE-9B). The example shows differences in detail, context, and narrative style across scenarios like sports, transport, toys, and portraits.
    }
    \label{fig:captioning_task}
    \vspace{-0.2cm}
\end{figure*}

Table~\ref{tab:diff_backbone_s2} presents an evaluation of the LLaVA-MORE models that analyzes the impact of increasing image resolution when using the CLIP and SigLIP2 visual backbones. To tackle this, we leverage the S$^2$ scheme~\cite{shi2024we}, a widely adopted method designed to enhance image resolution. Specifically, we interpolate the input image to create additional copies with $2\times$ and $3\times$ the standard resolution accepted by the visual encoder. The copies are chunked into 4 and 9 squared images respectively, of the same resolution as the original one. In total, 14 images are generated per sample, each processed independently by the visual encoder. The resulting visual tokens are spatially pooled and then channel-wise concatenated, resulting in the same number of visual tokens, but with $3\times$ more channels. 

Notably, from Table~\ref{tab:diff_backbone_s2} (top), it can be seen that S$^2$ generally improves LLaVA-MORE-3.8B when using CLIP as the visual backbone. The improvements are even more consistent when switching to SigLIP2, which already works at a higher resolution than CLIP. However, when scaling up to \ours-9B (Table~\ref{tab:diff_backbone_s2} bottom), the benefits of S$^2$ vanish for some benchmarks. For instance, on Science-QA and AI2D, S$^2$ appears detrimental at the 9B scale, while it is advantageous with \ours-3.8B.

From these results, we can conclude that small-scale MLLMs may greatly benefit from working with high-resolution images. However, the positive impact of higher resolution appears to diminish as model size increases, and, ultimately, the benefits of increasing image resolution seem to be highly task-dependent.  
For instance, with TextVQA, a benchmark that heavily relies on detecting and recognizing text within images, S$^2$ consistently brings improvements. A similar pattern is found on GQA, which challenges MLLMs with questions requiring in-depth scene understanding. Curiously, we find substantial gains by applying S$^2$ on the Chinese version of MMBench (MMB-Cn), especially at the 3.8B scale, while its behavior on the English version
is conflicting: S$^2$ improves with SigLIP2, but it is detrimental 
with CLIP.

\vspace{0.21cm}
\begin{idea_4}
\centering
Input resolution and number of visual tokens are critical to achieving good results in multimodal benchmarks, especially with small-scale models.
\end{idea_4}

\subsection{Dataset Contribution During Pre-Training }
Finally, we analyze the effect of using different data sources for the pre-training phase. Specifically, we select 558K samples from three different sources for image-caption pairs: (i) samples directly derived from the LAION dataset, (ii) samples from Recap-DataComp-1B~\cite{li2024if} where captions are generated by an MLLM and provide a dense description of the scene, and (iii) a balanced combination of the first two configurations (\ie, LAION+Recap). Table~\ref{tab:diff_dataset} compares these three new datasets against the original LLaVA pre-training recipe (\ie, LLaVA Pre-Train LCS), consisting of a mixture of image-caption pairs from LAION~\cite{schuhmann2022laion}, CC3M~\cite{changpinyo2021conceptual}, and SBU~\cite{ordonez2011im2text}. 

Interestingly, \ours-3.8B never achieves the top score with the original mixture. Conversely, exclusively training on LAION stands out as the winning choice in 8 out of 13 benchmarks. This may be because LAION’s noisy web-sourced pairs closely resemble the nature of the training data used by the SigLIP2 visual encoder behind \ours-3.8B. In contrast, LLaVA-MORE-9B shows more robust performance across the different configurations. Notably, adding Recap samples boosts \ours-9B’s Chinese fluency, leading to the best score of 69.4 on MMB-Cn and improving other MLLM results.

\begin{idea_5}
\centering
The source of pre-training data plays a role at small scales, but it does not significantly impact medium-scale models.
\end{idea_5}

\subsection{Qualitative Results}
To better understand the behavior of different MLLMs beyond their quantitative performance on standard MLLM and VQA benchmarks, we also present qualitative results on the image captioning task\footnote{For these qualitative results, we use the prompt: ``\texttt{\small Describe this image.}''.}, as shown in Fig.~\ref{fig:captioning_task}.
In particular, we compare LLaVA-1.5-7B~\cite{liu2024improved}, which serves as our baseline, against our model based on Phi-4-3.8B and Gemma-2-9B using SigLip2 as visual encoder. As it can be seen, \ours versions can effectively describe input images and provide detailed and rich textual descriptions. For example, only the two versions of \ours successfully avoid hallucinations and provide an accurate description of the scene. In contrast, the LLaVA-1.5-7B model incorrectly identifies the object in question (\ie, erroneously recognizing the snowboard in the image as skis) highlighting a notable failure in visual understanding and alignment with textual output.

\begin{idea_6}
\centering
No single model configuration consistently outperforms others across all experiments. Performance highly depends on the specific task.
\end{idea_6}
\section{Conclusion}
This work presents a quantitative analysis of the integration of different LLMs and visual backbones into the LLaVA architecture, aiming to systematically and comparably evaluate the contribution of each component.
To the best of our knowledge, this is the first comparative analysis of different backbones conducted under consistent experimental settings, using the same dataset across all training stages and the same evaluation protocol. 
Moreover, this work also aims to develop a framework for training and evaluating different versions of MLLM. To support this, we will release the code along with the model checkpoints to encourage the community to experiment with new configurations.

\section*{Limitations}
 In this work, an analysis of a subset of publicly available LLMs and visual backbones was considered. The rapid evolution of this research area means that the results presented provide a snapshot of the best configurations currently achievable. In addition, preliminary results are presented on the integration of text-only reasoning models into multimodal tasks through fine-tuning. These insights can serve as a starting point to bridge the gap between text-only and multimodal reasoning.

\clearpage

\section*{Acknowledgments}
We acknowledge the CINECA award under the ISCRA initiative, for the availability of high-performance computing resources. This work has been conducted under two research grants, one co-funded by Leonardo S.p.A. and the other co-funded by Altilia s.r.l., and supported by the PNRR-M4C2 project ``FAIR - Future Artificial Intelligence Research'' and by the PNRR project ``ITSERR - Italian Strengthening of Esfri RI Resilience'' (CUP B53C22001770006), both funded by the European Union - NextGenerationEU.

{
    \small
    \bibliographystyle{ieeenat_fullname}
    \bibliography{bibliography}
}

\end{document}